\documentclass[letterpaper, 10 pt, conference]{ieeeconf}  

\IEEEoverridecommandlockouts                              

\usepackage{comment}
\usepackage{hyperref}
\usepackage{caption,graphicx,xspace}
\usepackage{subcaption,tikz}
\usetikzlibrary{positioning}

\title{\LARGE \bf
Lattices of sensors reconsidered when less information is preferred
}

\author{Yulin Zhang and Dylan A. Shell
\thanks{*This work was supported by NSF awards 
\href{http://nsf.gov/awardsearch/showAward?AWD_ID=1453652}{IIS-1453652},
\href{http://nsf.gov/awardsearch/showAward?AWD_ID=1527436}{IIS-1527436}
and \href{http://nsf.gov/awardsearch/showAward?AWD_ID=2034097}{IIS-2034097}.
Yulin Zhang and Dylan A. Shell are with the Dept. of Computer Science,
Texas A\&M University, College Station, TX, USA.
{\tt\small yulinzhang,dshell@tamu.edu}}%
}

\providecommand{\subsumes}{subsumes\xspace}

\providecommand{\proceeds}{\ensuremath{\preceq}}
\begin{document}

\maketitle
\thispagestyle{empty}
\pagestyle{empty}

\begin{abstract}
To treat sensing limitations (with uncertainty in both conflation of information and noise) we model sensors as covers. This leads to a semilattice organization of abstract sensors that is appropriate even when additional information is problematic (e.g., for tasks involving privacy considerations).
\end{abstract}
\section{Introduction}

Typically a robot designer pieces together sensors and other hardware to construct a
device capable of accomplishing some given task.  Choosing one sensor over
another requires understanding the information it can provide and capabilities
it can enable, as well as the various costs (to purchase, to power, to dispose of,
etc.) involved.  Roboticists have found that partial order relations are a
useful way to organize sets of sensors: LaValle~\cite{lavalle2019sensor} represents a
sensor as a partition on raw world features.  He uses the idea of one partition
being a refinement of another to give a notion of dominance and, thereby,
constructs a sensor lattice based on this relation.  
Some years earlier, O'Kane~\cite{OKa07}
introduced a relation between different models of robots by considering how
both sensing and actuation in one system might enable it to emulate another,
and using this, provided a partial order on the set of robot systems.
And since then Censi's work identified partially ordered set structures
on resources that implement some functionality~\cite{censi2015mathematical}. He showed that an antichain (a
non-comparable subset) characterizes the minimal set of resources\,---resources
including sensing accuracy---\,that suffice for some specific requirement.

This short note, having much in common with these antecedent works, offers
extensions and generalizations to the same line of inquiry.  It models
sensors in terms of the information that they provide and considers a few
different relationships between pairs of sensors. It offers an 
organization of the set of conceivable sensors, paying particular attention to their
role in discriminating states of the world needed to ensure that the robot can reach some goal.
As computational representations, the elements we describe have some utility as data-structure for sets of sensors to enable search for sensors and plans jointly. 

\begin{figure}[h]
\centering
\includegraphics[width=0.915\linewidth]{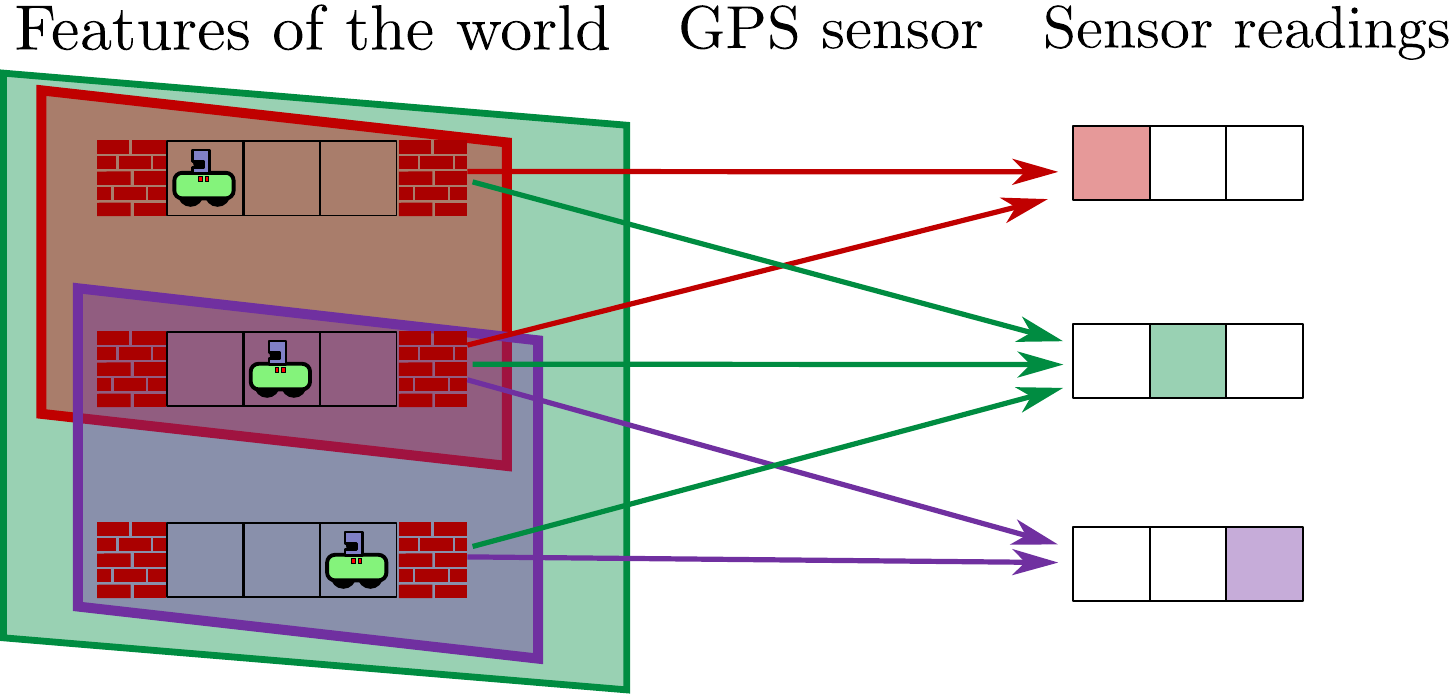}
\caption{A robot in a 3 state world has an imperfect GPS sensor that gives the true pose or a value off-by-one, non-deterministically. 
It is a many-to-many mapping from position (world feature) to sensor reading.
The pre-images of the GPS sensor form a cover. \label{fig:sensormap}}
\end{figure}



\section{Abstract sensors: Covers}

Sensors facilitate the flow of information from the world to the robot.  We
treat the world as having features of interest, and sensors as set-valued maps
providing the possible readings associated with those features. The robot's
task is to interpret a given reading, i.e., to find those aspects of the world
that correspond to the reading. This inverted use of the map means that we are
interested in the pre-images of elements in the sensor map's co-domain.  The
illustration in Figure~\ref{fig:sensormap} helps depict how this gives rise to
a cover on the world features: that is to say, a collection of sets. Real sensing is often said to be \emph{limited.} Two
different aspects of ambiguity can be observed directly in the sets
that comprise the cover: ($i$) the size of a set characterizes how much conflation
occurs owing the projective nature of a sensor (the world is complex, its
features are many, but sensor outputs are comparatively few); ($ii$) the degree of
overlap indicates aspects `noise' (whether it be non-deterministic or stochastic).
We will require that a sensor always output \textsl{some} value. 
So, in this model,  covers that correspond to sensors must ensure all world features are covered.

\section{Lattices, Sensors, and Sensor Lattices}

\begin{figure}[ht!]
\centering
\resizebox{0.5\textwidth}{!}{
\begin{tikzpicture}[scale=.7, every node/.style={font=\tiny}]
\node (r11)  at (-1, 2) {$\lbrace \lbrace 1, 2, 3\rbrace \rbrace$};
\node[below left=0.3cm and 2.0cm of r11] (r21) {$\lbrace \lbrace 1\rbrace, \lbrace 1, 2,3\rbrace \rbrace$};
\node[right=0.01cm of r21] (r22) {$\lbrace \lbrace 1,3\rbrace, \lbrace 1, 2,3\rbrace \rbrace$};
\node[right=0.55cm of r22] (r23) {\small \dots};
\node[right=0.7cm of r23] (r26) {$\lbrace \lbrace 1,2\rbrace, \lbrace 2,3\rbrace \rbrace$};
\node[below =0.3cm of r21] (r31) {$\lbrace \lbrace 1\rbrace, \lbrace 2\rbrace, \lbrace 1, 2,3\rbrace \rbrace$};
\node[right=0.01cm of r31] (r32) {$\lbrace \lbrace 3\rbrace, \lbrace 1,3\rbrace, \lbrace 1, 2,3\rbrace \rbrace$};
\node[right=0.01cm of r32] (r34) {\small \dots};
\node[right=0.25cm of r34] (r35) {$\lbrace \lbrace 1\rbrace, \lbrace 1,2\rbrace, \lbrace 2,3\rbrace \rbrace$};
\node[below =0.3cm of r31.south west, anchor=north west] (r41) {$\lbrace \lbrace 1\rbrace, \lbrace 2\rbrace, \lbrace 3\rbrace, \lbrace 1, 2\rbrace, \lbrace 2,3\rbrace, \lbrace 1, 2,3\rbrace \rbrace$};
\node[right=0.01cm of r41] (r42) {\small \dots};
\node[right=0.01cm of r42] (r43) {$\lbrace \lbrace 1\rbrace, \lbrace 2\rbrace, \lbrace 3\rbrace, \lbrace 1, 2\rbrace, \lbrace 2,3\rbrace, \lbrace 1, 3\rbrace \rbrace$};
\node[below right =0.3cm and 2cm of r41.south west, anchor=north west] (r51) {$\lbrace \lbrace
1\rbrace, \lbrace 2\rbrace, \lbrace 3\rbrace, \lbrace 1, 2\rbrace, \lbrace
2,3\rbrace, \lbrace 1, 3\rbrace \rbrace, \lbrace 1, 2, 3\rbrace \rbrace$};
\draw (r11) -- (r21) -- (r31); 
\draw (r41) -- (r51);
\draw[dashed] (r31)--(r41);
\draw (r11) -- (r22) -- (r32); 
\draw (r11) -- (r23);
\draw[dashed] (r34)--(r42);
\draw (r43) -- (r51); 
\draw (r26) -- (r35); 
\draw[dashed] (r32) -- (r41); 
\draw (r23) -- (r34); 
\draw[dashed] (r35) -- (r43); 
\draw (r42) -- (r51); 
\end{tikzpicture}
}
\caption{The cover semilattice for a trivial three-element set. \label{fig:lattice}}
\end{figure}
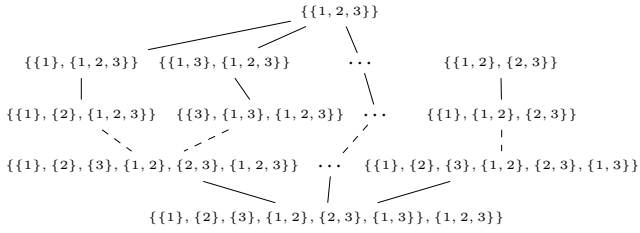

Given two covers, the subset relation gives a straightforward property with which to compare them. 
If cover $A$ is a sub-collection of the pre-images appearing in $B$, i.e., $A \subseteq B$, then we say that $A$ \emph{\subsumes} $B$. If $A$ \subsumes $B$, then we place $A$ above $B$ in a diagram, and hence $B$ is a lower bound of $A$. For any two covers $A$ and $B$, there always exists a smallest cover that contains all pre-images of $A$ and $B$. 
This smallest cover (namely $A \cup B$) is a greatest lower bound (``greatest'' means that the number of pre-images is as small as possible). 
In contrast, for any two covers that contain different pre-images,
there may not exist any least upper bound for them
(their intersection leaves some features uncovered). Hence, the covers organized via the subsumption (or subset) relation form a meet-semilattice. An example of a semilattice for a trivial three world features is shown in Figure~\ref{fig:lattice}.

The robot in a partially observable environment maintains an estimate of possible world states.
Under worst-case uncertainty, sensing information shrinks the set possible world states in the estimate.
Successful execution demands that the robot choose some progress-making action that is safe on all estimated world states. 
Suppose a robot can reach its goal from  one junction only by turning left, and  from another junction only by turning right.
Selecting a progress making action is infeasible when these two junctions are indistinguishable to the robot.
Sensor information that helps to distinguish the two can make the problem solvable. 


Owing to noise, the robot may non-deterministically receive one of  multiple possible sensor readings (in the image set of the same world feature).  
Suppose a robot has a plan which is guaranteed to reach  a goal region. 
In some place, it must have have actions that make progress for any sensor readings that it might receive.
As a consequence, if the goal is attainable under a sensor cover $A$, then the goal is also attainable under any sub-collection of $A$ that covers all the world features.
If plan $P$ for $A$ reaches the goal region, and $B$ \subsumes $A$, then
$P$ also works for~$B$.

%

Let $(C)^{u}$ denote the set of all valid covers that are subsumed by $C$, and call this the \emph{u-inflation} of $C$.
Given a set of covers, the \emph{upper covers} are the subset of covers that do not subsume any other cover in the set.
There may be multiple upper covers for a set, within the subsumption semi-lattice they lower bound the set.
As shown in~\cite{Zhang2020},
they can be used to search for sensors which provide adequate information for planning problems to be soluble---since the existence of a goal-attaining plan for an upper cover implies the same for any covers that subsume it, i.e., all those generated via u-inflation.
Upper covers and the preceding statements have considered the number of readings a sensor produces together with the role of non-determinism in (worst-case) planning; next, we dig a layer deeper to look not at the cardinality of the cover, but the contents of the sets comprising it.

Smaller pre-images represent finer sensor readings, and
finer readings better discriminate world features. They yield tighter beliefs, which may be vital for plans to ensure that they reach goal regions.
But this capacity is determined not by the finest sensor reading but the coarsest one, since the robot must be able to reach the goal under all readings it may receive.
Including additional sensor readings that are finer than others already in a some cover, will have no effect on that sensor's capability  with regards to goal attainment.
Let $V^{\star}$, the \emph{star-closure}, be the cover obtained by adding new, finer sensor readings to $V$, until a fixed point is reached.
We say two covers $A$ and $B$ are star-equivalent, denoted \mbox{$A\sim_{\star} B$}, if and only if $A^{\star}=B^{\star}$. 
Note that two covers with one subsuming the other may also be star-equivalent (e.g., take
$\lbrace \lbrace 1,2\rbrace \rbrace$ and $\lbrace \lbrace 1,2\rbrace, \lbrace 1\rbrace \rbrace$), and thus
quotienting by $\sim_{\star}$ collapses elements from multiple layers in the cover semilattice.
This yields a second partial order relation, \emph{star-subsumption}: 
two covers can be ordered by the subsumption of their star-closures.  It turns out that the upper cover concepts, and statements made earlier, are applicable on the quotient semilattice. 

A richer partial order that combines subsumption and star-subsumption can given:  For covers $A$ and $B$, we say $A\proceeds B$, if ($i$)~$A\subseteq B$ or ($ii$)~$A\not\subseteq B$, $B\not\subseteq A$, and $A^{\star}\subseteq B^{\star}$. 
But why care about the plain subsumption relation then? 

There can be value to a robot maintaining ignorance~\cite{OKa08,eick2020enhancing}; some 
tasks may stipulate that certain information is
not to be learned (by the robot, or an observer~\cite{zhang18complete,zhang18planning}).
Sensors that collect more information are not always to be preferred, but the star-closure assumes that it does no harm.
When divulging too much information is problematic, one must
consider sensors on the original cover semilattice, as the
quotient semilattice will fail to be a `congruence' with respect to questions of solubility (some elements in $[A]_{\sim_\star}$ might meet the stipulations, while others fail to).

\subsection{LaValle's partition sensor lattice revisited}
Partitions are special covers. 
For any two partitions, $Q_1$ and $Q_2$, neither subsumes
the other. Using the plain subsumption relation, one 
obtains a collection of mutually incomparable elements, they form a degenerate lattice. 
When more information is better then the star-closure is useful and a star-subsumption ordering (or $\proceeds$) will make sense. 
If the cover happens to be a partition, then the star-closure of a finer partition subsumes the star-closure of a coarser partition. 
Thus, one recovers LaValle's partition sensor lattice~\cite{lavalle2019sensor} in its entirety as a slice of 
the cover semilattice ordered by $\proceeds$. Also, 
picking the sets with minimal cardinality in star-equivalent classes to be representatives, 
LaValle's lattice also appears within the $\sim_\star$ quotient
semilattice.

\section{Coda}

This paper studies partial order relations on abstract sensors in terms of both the sets of readings they generate and the discriminating power they provide. 
The sensors form a semilattice structure.
Planning to reach goal states subject to worst-case non-determinism gives a particular interpretation for the information obtained from the sensors. 
These semantics allow for collections of sensors
to be condensed and described through a lower bounding set.
When one may assume that greater discriminating fineness is never detrimental, further reductions can also be made via an equivalence relation. 
Despite the words `lower bounding' suggesting otherwise,  the former remains valuable even for tasks where more information is not always better.
Finally, an ordering relation ($\proceeds$) treats both of these aspects together and generalizes LaValle's sensor lattice.

\newpage
\IEEEpeerreviewmaketitle

\bibliographystyle{IEEEtran}
\bibliography{mybib}

\end{document}